# RGB-T Tracking Based on Mixed Attention

Yang Luo[1, 2], Xiqing Guo[1, 2*], Mingtao Dong[3], Jin Yu[1, 2]

luoyang211@mails.ucas.ac.cn, guoxq100036@aircas.ac.cn, mingtao_d@163.com, jinyu@aoe.ac.cn

*Abstract*—**RGB-T tracking involves the use of images from both visible and thermal modalities. The primary objective is to adaptively leverage the relatively dominant modality in varying conditions to achieve more robust tracking compared to single-modality tracking. An RGB-T tracker based on mixed attention mechanism to achieve complementary fusion of modalities (referred to as MACFT) is proposed in this paper. In the feature extraction stage, we utilize different transformer backbone branches to extract specific and shared information from different modalities. By performing mixed attention operations in the backbone to enable information interaction and self-enhancement between the template and search images, it constructs a robust feature representation that better understands the high-level semantic features of the target. Then, in the feature fusion stage, a modality-adaptive fusion is achieved through a mixed attention-based modality fusion network, which suppresses the low-quality modality noise while enhancing the information of the dominant modality. Evaluation on multiple RGB-T public datasets demonstrates that our proposed tracker outperforms other RGB-T trackers on general evaluation metrics while also being able to adapt to long-term tracking scenarios.**

## I. INTRODUCTION

Visual object tracking is an essential branch of computer vision that has recently made significant progress and has been widely applied in various industries, including autonomous driving, intelligent security, and robotics [1]. The current research on visual object tracking predominantly focuses on visible modality (RGB tracking). This is primarily due to the rich colour and texture information that RGB images provide, making target feature extraction more feasible, and relatively low-cost imaging devices based on visible modality [2]. However, researchers have identified the limitations of RGB tracking in extreme environments such as rain, fog, and low-light conditions. Thermal infrared imaging, on the other hand, relies on the target's thermal radiation and is less affected by intensity variations in illumination, making it possible to penetrate rain and fog. It provides a complementary and extended modality to RGB tracking [3]. This complementarity has led researchers to progressively concentrate more on object tracking based on the fusion of these two modalities to create trackers with higher accuracy and robustness [4].

Object tracking based on the fusion of RGB and thermal images is also known as RGB-T tracking. Most existing RGB-T tracking methods employ convolutional neural networks (CNN) to extract target features by down-sampling the image and expanding the perceptual field. Subsequently, the weights of different modality feature maps are computed to achieve feature fusion.

However, some researchers have indicated that CNNs may not effectively learn long-range pixel dependencies [5]. Despite continuous down-sampling, the actual growth rate of CNN's receptive field can be much slower than expected. This results in a significant amount of local information being retained, making it difficult for CNNs to consider both the target template and background contextual information, which is essential for object tracking. In contrast, transformer networks based on the attention mechanism do not have these limitations and do not require stacking network layers to obtain global information [6]. Fig. 1 illustrates the disadvantage of CNN-based models (APFNet[1], TFNet[7], MDNet[8], MA-Net++[9]) in obtaining global information in RGB-T tracking. Specifically, the CNN-based models tend to maintain the size and shape of the bounding box in the initial template frame and struggle to distinguish target instance features. As a result, they still use local information to track the target, making it difficult to achieve accurate results when the target undergoes significant deformation and displacement. The proposed method (MACFT) in this paper overcomes these shortcomings by using transformer as the feature extraction backbone. By examining global information, our model can better "understand" the target, accurately locate it even after significant deformation, and improve tracking performance.

In addition to feature extraction from RGB and thermal images, achieving complementary fusion of both modalities is essential for RGB-T tracking.

Earlier research such as MANet proposed by Li et al. [10] only simply concatenated(Concat) the features from the RGB images and the thermal images before feeding into the fully connected network layer for learning. On the other hand, Zhang et al. [8] designed a modality weight calculation structure for search image features based on the Siamese dual-stream network, but for template image features, still simply fused using Concat. Zhu et al. [11] designed a structure based on the attention-based feature fusion proposed in SKNet [12], which adaptively modulates the neural receptive fields in the modality fusion stage. Xiao et al. [1] used a similar structure to fuse challenging image sequences for targeted learning. Furthermore, a stacked structure of encoder and decoder transformer layers that improve the interaction of information between modalities has achieved excellent results on multiple datasets. The limitation of this method is that it still relies on CNN for generic feature extraction and performs attention operations in the high-level semantic space after multiple downsampling, ignoring finer-grained image information.

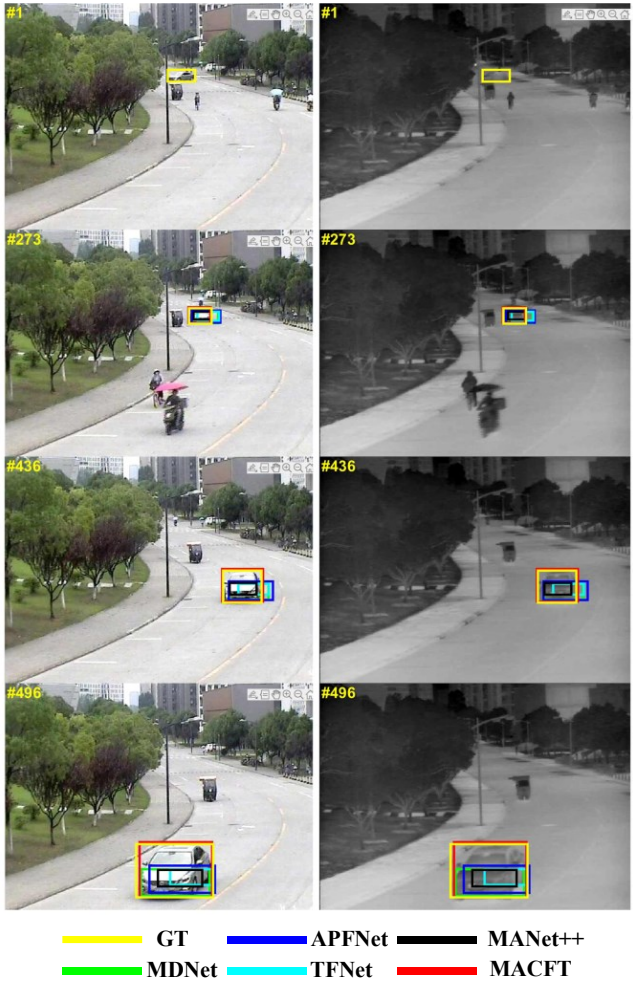

**Fig. 1.** Comparing the model proposed in this paper with other advanced CNN-based RGB-T tracking models when the target undergoes large deformation(The yellow box represents ground truth). We can find that our MACFT model better "understands" the characteristics of the target and generates a bounding box that fits more closely to ground truth.

To enhance our algorithm's "understanding" of the tracking target and and enable adaptive fusion using complementary information from different modalities , we propose a novel RGB-T tracking method (referred to as MACFT) based on mixed attention mechanism. The modalities' specific features are extracted from the RGB and thermal images first using a two-branch transformer backbone, and then the modalities' shared features are extracted using a transformer backbone with shared parameters. The features of each branch are fused using a fusion network based on mixed attention mechanism and passed to a corner predictor network to output the predicted bounding box.

In the feature extraction stage, we used the pre-trained transformer model provided in [13] as our backbone. During training, a mixed attention operation is constructed by feeding the concatenated template and search images into the transformer backbone. The self-attention operation is used to extract features of the template and search images themselves, while the cross-attention operation enables sufficient infor-mation interaction between the target template and the search images. This allows the model to better understand the semantic features of the target based on the template and search images' contextual information. On this basis, it is considered that the RGB and thermal images, although having different imaging bands, still have many correlations, such as the target's edge and contour information and some fine-grained texture information [9]. Therefore, we introduce an additional shared-parameter branch alongside the two-branch network to learn features shared between modalities by introducing a KL divergence loss function to constrain the consistency between modality features.

In the modality feature fusion stage, inspired by [14], we propose a structure for fusing complementary features between visible and thermal modalities. Its core is Mixed Attention Module (hereinafter referred to as MAM) and Cross Attention Module (hereinafter referred to as CAM). MAM can perform both self-attention and cross-attention operations on the features from the visible and thermal modalities, while CAM only includes cross-attention operations. The cross-attention operation facilitates cross-modal information interaction, while the self-attention operation enhances the fused information from both modalities.

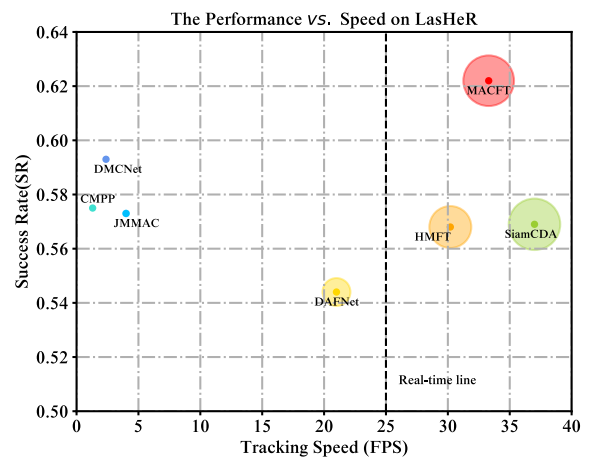

**Fig. 2.** Comprehensive comparison of tracking success rate and speed of LasHeR dataset. Our MACFT model achieves the highest tracking success rate under the premise of ensuring real-time running(33.3 FPS on average).

After cross-fusing features from the modality-specific feature branch and the modal-shared feature branch, we generate the target bounding box with only a simple structured corner point predictor. After extensive experiments on several public RGB-T tracking datasets, we verify that the proposed method is effective and outperforms most of the current state-of-the-art RGB-T tracking models. Our method runs at 33.3 FPS (50.2 FPS for a lighter version with modal-shared feature branches removed) on an RTX 3090 GPU and can run in real-time. Fig. 2 gives a visual display of the comprehensive comparison of MACFT's success rate and running speed on the LasHeR dataset with other advanced RGB-T tracking methods (the compared methods include CMPP [15], DMCNet [16], JMMAC [17], HMFT [18], SiamCDA [19], and DAFNet [20]).

The main contributions of our work can be summarized as follows:

- We propose a high-performance RGB-T fusion tracking framework named MACFT, which addresses the extraction of shared and specific features of visible and thermal modalities. By introducing mixed attention operations in the backbone, we construct a robust target feature representation with strong adaptability.

- We designed a feature fusion structure based on mixed attention mechanism and cross-attention mechanism, which cross-fuses and enhances the features from different backbone branches. This effectively suppresses low-quality modality information and enhances advantageous modality information in an adaptive manner, leading to more effective information fusion. Moreover, it can achieve remarkable performance improvements using only a small number of trainable parameters and training epochs.

- After extensive testing and evaluation, our proposed method can achieve state-of-the-art performance on several challenging datasets. Including RGBT234 [21], LasHeR [22], and VTUAV [18].

## II. Related Work

### A. RGB tracking methods

Visual object tracking based on RGB images has been developed for quite some time, but it wasn't until recent years when deep neural networks were successfully applied to computer vision tasks that models with high tracking accuracy emerged. They can be broadly classified into two categories, namely discriminative models and generative models.

Discriminative models are trained using the whole image labeled with the target location, aiming to obtain a classifier that can distinguish between foreground and background, and often need to generate multiple proposal regions during the tracking process, then feed the classifier through the forward propagation to get positive and negative samples. These samples are further refined through post-processing methods such as cosine window penalty, and the best samples are retained. Typical discriminative trackers such as MDNet [8], ATOM [23], DiMP [24], etc. Although these models are known for their high accuracy, the excessive online processing required for them consumes many computational resources, which results in slow operation. To address this drawback, Jung et al. [25] modified the forward propagation based on MDNet with the ROI Align [26], thereby increasing its speed.

Contrary to the discriminative model, the generative model solely relies on offline training and has no online update step. The final prediction of the target is obtained by computing the joint probability density of both the template and search image. Typical generative models such as the Siamese network-based tracker [27]–[29]. The generative model's major strength is its small computational effort and fast speed, but its tracking accuracy is often inferior to that of discriminative models due to insufficient information interaction between the template and the search images and the lack of an online up-date step.

In recent years, the field of computer vision has witnessed the gradual replacement of CNN as the backbone architecture with the transformer. In the domain of visual object tracking, Chen et al. [30] designed self-contextual augmentation (ECA) and cross-feature augmentation (CFA) modules in their proposed model TransT that differ from the traditional transformer structure to achieve a better fusion of features between templates and search regions. Yan et al. [32] proposed a method based on the transformer codec structure to achieve tracking directly by bounding box prediction, in which the image is fed into a simple fully convolutional network for feature extraction and then directly fed into the transformer prediction head for bounding box prediction without pre-generating proposal regions. Cui et al. [14] proposed an asymmetric attention mechanism for information interaction between template and search images, which saves computational resources and provides the possibility to update template online during the tracking process.

To improve the speed of our tracker, the model proposed in this paper is based on the idea of the generative model, the process of tracking relies entirely on offline training. Inspired by the work of Chen et al. [13], the full interaction between the target template and the search image is implemented within the transformer backbone, which makes our tracker a very high accuracy as well.

### B. RGB-T tracking methods

RGB-T tracking is a subset of visual object tracking in which the key is to effectively use complementary information from multiple modalities to achieve a more stable and robust tracking results. In 2011, Wu et al. [31] proposed an RGB-T tracking model based on sparse representation, leading to the rapid development of RGB-T tracking, and similar works were proposed by [32]–[34], etc. Li et al. [35] proposed a graph descriptor that automatically learns global and local features to achieve the fusion of multi-modal information by imposing low-rank constraints on the joint parameter matrix of multi-modal images. In addition, Li et al. [36], [37]also employed a two-stage regularized ranking algorithm to optimize image patch weights and modality weights for obtaining robust features while suppressing background noise. Further, Wang et al [38] introduced correlation filter to RGB-T tracking for the first time in 2018, using the average correlation peak energy to calculate the weights. Recently, Zhang et al. [17] proposed a robust RGB-T tracking framework called JMMAC, which takes ECO [39], a well-known algorithm in single-modality tracking, as the baseline and takes motion principles into full consideration to design an adaptive structure that allows the model to switch between appearance modeling and motion modeling.

In recent years, more high-performance RGB-T trackers have been proposed because of the introduction of deep neural networks. Li et al. [10] developed a multi-adapter convolutional network for RGBT tracking by following MDNet's multi-domain learning framework that performs feature extraction for modal-generic features, modal-specific features, and in-stance features. Subsequently, Lu et al. [9] improved their work by applying a hierarchical divergence loss function and ROIAlign with reference to RT-MDNet [25] to reduce the number of forward propagations and enhance real-time performance. Zhu et al. [40] also used MDNet, but through a 1 × 1 convolutional kernel + Relu + LRN (local response normalization) to form a feature aggregation module to fuse the features of two modalities and randomly reject the feature channels by the channel dropout technique, which effectively eliminates redundant information and noise while preventing model overfitting. Zhang et al. [41] first implemented the task of RGB-T fusion tracking based on Siamese network combined with multi-modal features. In this method, the modality weight calculation is designed according to the size of the predicted target position deviation, which improves the reliability of the model.

With the increasing trend of transformer architecture surpassing CNN in terms of performance in recent years, some researchers in the field of RGB-T tracking have also tried to introduce the core idea of it, the attention mechanism, into their models. A typical example is APFNet proposed by Xiao et al. [1] which adopts a specific strategy for fusion based on RGBT-specific attribute challenges in feature extraction. To achieve modality self-enhancement and inter-modality interaction enhancement, they incorporated three independently separated encoders (based on self-attention mechanism) and decoders (based on cross-attention mechanism) in each fusion branch. However, there are few RGBT tracking models that base their feature extraction entirely on the transformer backbone.

### C. Multi-modal image fusion methods

How to learn and fuse the complementary information between different modalities to adaptively change the weights corresponding to a modality when the quality of a modality decreases is one of the key issues of RGB-T tracking.

According to the difference in the levels of image fusion of different modalities, RGB-T modality fusion algorithms can be classified as pixel-level, feature-level, and decision-level. Pixel-level fusion mainly performs shallow fusion of images, fusing complementary information between modalities on fine-grained features of the image. Chaib et al. [42] proposed a fusion method for VHR remote sensing scene classification, and Shao et al. [43] used local hold projection for image fusion and applied it to rotating machinery fault diagnosis. However, the drawback of pixel-level fusion is that it is computationally intensive and not conducive to online inference. Most of the current deep neural network-based RGB-T trackers use a feature-level and decision-level fusion approach, where features are first extracted through a backbone, and then features from different modalities are fused at the middle or rear of the model. Guo et al. [44] extended the SiamRPN [27] framework to a dual Siamese subnetwork for RGB-T tracking and introduced the weight distribution of the two modal features right at the feature extraction stage and designed a joint modality channel attention module to assign the corresponding weights to the channels of the depth features. Zhang et al. [18] based on the mfDiMP [45] framework, unified multi-modal fusion strategies (including pixel-level fusion, feature-level fusion, and decision-level fusion) into a hierarchical fusion framework and designed a self-attention module to model modality confidence and non-locally explore modality information.

The advantage of the feature-level and decision-level fusion approaches is that the computational effort is relatively small, which is beneficial to ensure the real-time performance of the algorithm. In our proposed model, a mixed attention-based structure is used to integrate the features from each backbone branch in a unified way, which is feature-level fusion.

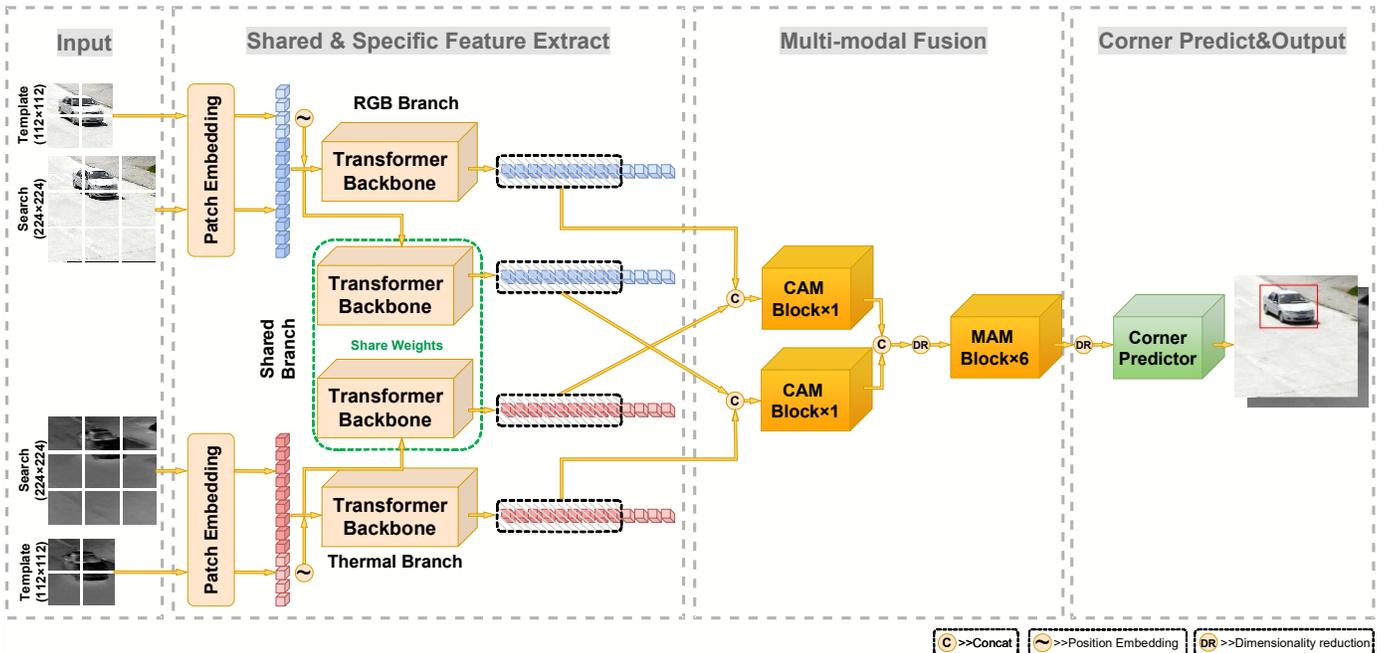

**Fig. 3.** An overview of the MACFT model, which is divided into three parts, which are used for modal-specific/shared feature extraction, information fusion between modalities, and bounding box regression.

## III. METHOD

In this section, our proposed RGB-T tracking model MACFT is described in detail, as shown in Fig. 3. The whole model consists of three parts, which are the modality complementary feature extraction network based on transformer backbone, the modality adaptive fusion network based on mixed attention mechanism, and the target localization regression network. Each of these components will be discussed specifically in the following.

### A. Transformer-based complementary feature extraction backbone

In our proposed model, the feature extraction part is divided into three branches, namely, the visible modal-specific feature extraction branch, the thermal modal-specific feature extraction branch, and the modal-shared feature extraction branch. ViT [6] served as the base backbone for all three branches, initilized with parameters from CLIP [46] pre-trained model. Furthermore, the template image is concatenated with the search image in the same dimension to achieve information interaction. The process is detailed as follows.

#### 1) Model-specific feature extraction branch

First, given a video sequence, we select the first frame as the reference frame and crop the template image $\mathbf{Z} \in \mathbb{R}^{H_z \times W_z \times 3}$ according to the labeled bounding box, where $H_z \times W_z$ is the size of the template image, and accordingly, the subsequent frames in the video sequence are used as the search image with the size set to $\mathbf{X} \in \mathbb{R}^{H_x \times W_x \times 3}$. To accommodate the generic ViT backbone, we set the search image size $H_x \times W_x$ to 224×224 and the template image size $H_z \times W_z$ to 112×112.

Before using the transformer backbone for feature extraction, the input image is first serialized, i.e., patch embedding, by dividing the 2D image into N blocks of size $D = P^2 \times C$ ($P$ is the size of patch, here $P$=16 since the backbone used is ViT-B-16, and $C$ is the number of channels of the image(here $C$=3), then project the patch onto the space with dimension $D$ through a linear transformation. For our input template image and search image, the dimensions of the image after serialization will be transformed into $\mathbf{Z}_p \in \mathbb{R}^{N_z \times D}$ and $\mathbf{X}_p \in \mathbb{R}^{N_x \times D}$, where $N_z = H_z W_z / P^2$ and $N_x = H_x W_x / P^2$.

For the transformer backbone, there is an additional step in the input process which is to add positional encoding to help the model distinguish the order of the input sequence. For the search image sequence, the pre-trained positional encoding $p_s = p_{ptr}$ can be used directly because its size is consistent with the pre-trained ViT backbone input. However, for the template image, its size is not align with the ViT input and cannot accommodate the pre-trained positional encoding, so we set up a learnable positional encoding structure that consists of a two-layer fully connected network, which is added to the template image sequence to obtain the positional encoding vector $p_z$ of the template image.

After adding the position encoding vector, we obtain the final input vectors $z \in \mathbb{R}^{N_z \times D^*}$ and $x \in \mathbb{R}^{N_x \times D^*}$, and subsequently, we concatenate $z, x$ along the first dimension and feed them into the transformer backbone network for learning.

Let the concatenated vector be $r$:

$$r = Concat(z, x, dim = 0). \tag{1}$$

Let the output of layer n in the backbone network be $r^n$, then we have:

$$[r^*] = [r^{n-1}] + MA([r^{n-1}]), \\ [r^n] = [r^*] + FFN([r^*]). \tag{2}$$

where $MA$ denotes mixed attention operation and $FFN$ denotes feed forward network. The calculation within $MA$ can be expressed as follows:

$$MA([r^n]) = \\ softmax\left(\begin{bmatrix} a(z^n, z^n), a(z^n, x^n) \\ a(x^n, z^n), a(x^n, x^n) \end{bmatrix}\right)\left(\begin{bmatrix} z^n W_V \\ x^n W_V \end{bmatrix}\right). \tag{3}$$

where $a(x, y) = (xW_Q)(yW_K)^T / \sqrt{d}$.

It is in the cross-attention operations such as $a(z^n, x^n)$ & $a(x^n, z^n)$ that the model learns the relationship between the template image and the search image, and in the self-attention operations like $a(z^n, z^n)$ & $a(x^n, x^n)$ that this relationship is self-enhanced. We construct two identical and independent feature extraction branches for RGB and thermal images based on the above method for extracting features unique between different modalities. Let the feature vectors output from the two branches be:

$$\mathbf{R}_v \in \mathbb{R}^{(N_z+N_x) \times D^*}, \mathbf{R}_t \in \mathbb{R}^{(N_z+N_x) \times D^*}. \tag{4}$$

#### 2) Modal-shared feature extraction branch

To better construct a complementary feature representation between the visible and thermal modalities, we add a modal-shared feature extraction backbone to the model (again using ViT as the backbone with the same input structure as the modal-specific feature extraction branch) for learning features shared by both modalities. Also, to be able to constrain the backbone of this branch to ensure that its output features are consistent across the two modalities, we introduce a Kullback-Leibler (KL) divergence loss $\mathcal{L}_{div}$. The formula is expressed as follows:

$$\mathcal{L}_{div} = \text{KL} \left(\mathbf{G}_v \| \mathbf{G}_t\right) \\ = \frac{1}{N_z+N_x} \sum_{n=1}^{N_z+N_x} (g_{vn} log \, (g_{vn} - g_{tn})). \tag{5}$$

where $\mathbf{G}_v \in \mathbb{R}^{(N_z+N_x) \times D^*}$, $\mathbf{G}_t \in \mathbb{R}^{(N_z+N_x) \times D^*}$ denote the output vectors (after *Softmax*) of the visible modality input and thermal modality input through the shared feature branch backbone network, respectively, while $g_{vn}$、$g_{tn}$ denote the n[th] term of the first dimension in $\mathbf{G}_v$、$\mathbf{G}_t$. KL divergence based on the information entropy principle, it can measure accurately the information lost when approximating one distribution with another, so it can be used to calculate the similarity between two distributions.

### B. Mixed attention modality fusion network

After obtaining the features from the three branches, the visible branch, the thermal branch, and the shared branch, we design a modality adaptive fusion network based on mixed attention mechanism and cross-attention mechanism, with the aim of learning an adaptive discriminative weight for features from different modalities, in order to enhance high-quality modality and suppress low-quality modality.

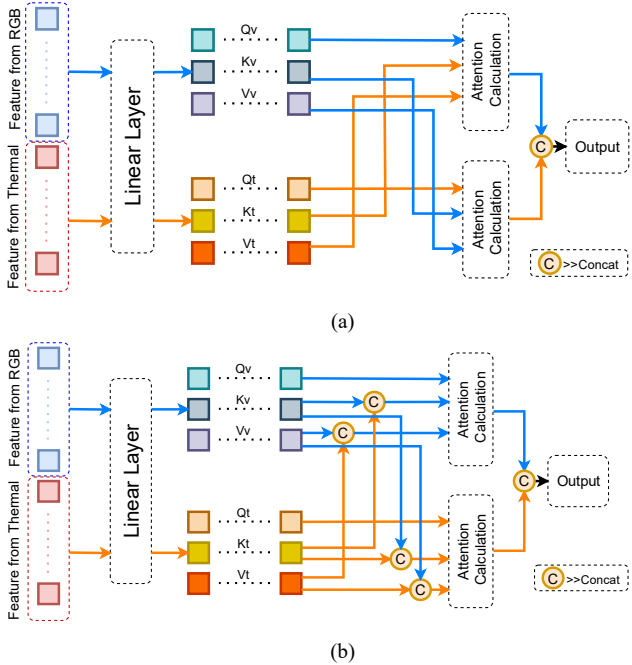

**(a)**

**(b)**

**Fig. 4.** Schematic diagram of the MACFT modality fusion network structure, (a) represents CAM, (b) represents MAM, CAM only includes cross-attention operations between RGB and thermal features, and MAM introduces self-attention based on CAM.

As shown in Fig. 4, we use two types of attention modules in the modality fusion network, the cross-attention module (CAM) and the mixed attention module (MAM).

Given a vector consisting of visible features connected with thermal features $x = Concat(x_v, x_t)$, we first map the input into three equally shaped weight matrices called *query*, *key*, and *value* by a linear layer. $q_v$, $k_v$, $v_v$ is used below to denote the visible part and $q_t$, $k_t$, $v_t$ denote the thermal part. From (3), the calculation of MAM can be described as:

$$MAM([x]) =$$
$$softmax\left(\begin{bmatrix} a(x_v, x_v), a(x_v, x_t) \\ a(x_t, x_v), a(x_t, x_t) \end{bmatrix}\right)\left(\begin{bmatrix} x_v v_v \\ x_t v_t \end{bmatrix}\right). \quad (6)$$

where $a(x_v, x_t) = (x_v q_v)(x_t k_t)^T / \sqrt{d}$, $a(x_v, x_v)$, $a(x_t, x_t)$, $a(x_t, x_v)$ in the same way.

The CAM module, on the other hand, eliminates the self-attention component and retains only the cross-attention, which can be described as:

$$CAM([x]) =$$
$$\begin{bmatrix} softmax(a(x_v, x_t)) * x_t v_t \\ softmax(a(x_t, x_v)) * x_v v_v \end{bmatrix}. \quad (7)$$

In the modality fusion network, we connect the outputs from the visible modal-specific feature branch with the modal-shared feature branch (thermal part), and from the thermal modal-specific feature branch with the modal-shared feature branch (visible part), but select only the part with the same size as the search image and discard the template part:

$$R_v, R_t \in \mathbb{R}^{(N_z+N_x) \times C} \rightarrow R_v^x, R_t^x \in \mathbb{R}^{N_x \times C},$$
$$G_v, G_t \in \mathbb{R}^{(N_z+N_x) \times C} \rightarrow G_v^x, G_t^x \in \mathbb{R}^{N_x \times C}. \quad (8)$$

$$H_{vt} = Concat(R_v^x, G_t^x),$$
$$H_{tv} = Concat(R_t^x, G_v^x). \quad (9)$$

Subsequently, $H_{vt}$, $H_{tv}$ are fed into the CAM module for modal information interaction:

$$C_{tv} = CAM(H_{tv}),$$
$$C_{vt} = CAM(H_{vt}). \quad (10)$$

Only cross-attention is used here for the fusion of features from different branches due to the fact that sufficient self-enhancement has been performed in the backbone for the features from each modality, and there is no need to add extra self-attention modules. Additionally, removing the self-attention operation can also reduce the model's parameter count. However, in the following MAM modules, we introduce self-attention operation because the initial interaction mapped features from different modalities into a new space, necessitating further enhancement of feature representation with the self-attention mechanism. These points will be supported in Chapter IV in the subsection on ablation studies.

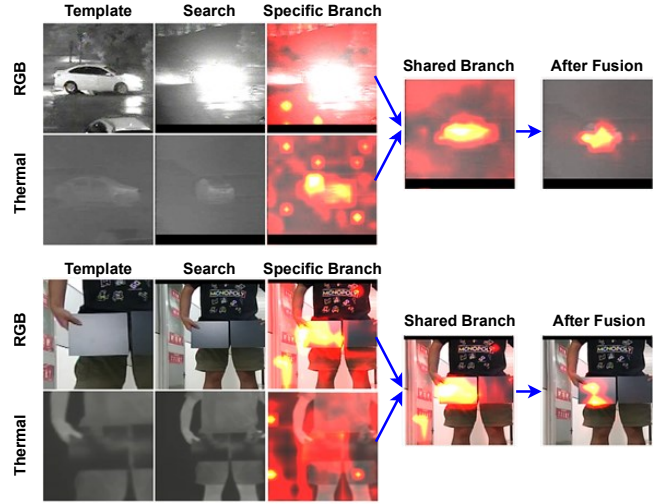

**Fig. 5.** Visualization of the attention weights of different branches and the attention weights after modality fusion network when one of the modalities is of low quality.

Finally, we connect the output of the CAM module and reduce the dimension and send it to the MAM module to further enhance the information interaction between the modalities:

$$C_{mix} = DR(Concat(C_{tv}, C_{vt})). \quad (11)$$

$$M_{mix} = MAM(C_{tv}, C_{vt}). \quad (12)$$

$$Output = DR(M_{mix}). \quad (13)$$

where $DR(\cdot)$ denotes the dimensionality reduction operation.

Experimentally, it is proved that the fusion structure we designed can well suppress the noise from low-quality modality while enhancing the feature expression of the dominant mo-dality, and the feature visualization is shown in Fig. 5.

We selected the carLight sequence (thermal modality is dominant) from the RGBT234 dataset and the leftmirror sequence (visible modality is dominant) from the LasHeR dataset for visualization. Among the attention weights maps, we can obviously find that the noise introduced by the low-quality modality images is well suppressed after our designed fusion structure, while the features of the dominant modality images are further refined.

### C. Target localization regression network

After the modality fusion network, we use a lightweight corner predictor to localize the target. First, the features are reshaped into 2D, then, for the top left and bottom right corners of the bounding box, the features are downsampled using 5 convolutional layers, and finally the corner coordinates are calculated using the *softArgmax* layer and are reverse mapped to the coordinates corresponding to the original image according to the scale of the image crop.

### D. Training Method

#### 1) Training in stages

We employ a staged approach for model training, which comprises three stages. The first two stages involve backbone training, requiring the modality fusion network to be removed while retaining only the corner predictor.

In the first stage, we freeze the modal-shared feature branch and load the pre-trained model to initialize the backbone of the modal-specific feature branches. We then train the backbone of the corresponding branches using RGB and thermal image data and save the model parameters.

In the second stage, the modal-specific feature branches are frozen, the modal-shared branch is unfrozen. During the forward pass, RGB and thermal images are repeatedly used to update the parameters of the backbone, and the model parameters are saved.

In particular, during the training of the feature extraction backbone in the first two stages, in order to reduce the number of trainable parameters of the network and considering the shared characteristics of the low-level features between images, we froze the first eight multi-head attention modules of the ViT backbone and only fine-tuned the last four.

In the third stage, all feature extraction backbone branches are frozen, and the models trained in the first two stages are loaded for initializing the backbone parameters. The modality fusion network is added for training and the model parameters are saved.

#### 2) Loss function

The model proposed in this paper uses a total of three loss functions in the offline training phase, which are $l_1$ loss, generalized GioU loss [47] and KL divergence loss, and the loss functions are combined in the first and third stages of model training in the following manner:

$$\mathcal{L} = \lambda_{giou} \mathcal{L}_{giou}(b_i, b_i^*) + \lambda_{\mathcal{L}_1} \mathcal{L}_1(b_i, b_i^*). \quad (14)$$

where $\mathcal{L}_1$ denotes the $l_1$ loss, $\mathcal{L}_{giou}$ denotes the generalized

GIoU loss, $b_i$ denotes the target bounding boxes predicted by the model on the search image, and $b_i^*$ denotes the true value of the target bounding boxes. And $\lambda_{giou}$ and $\lambda_{\mathcal{L}_1}$ are the weight parameters of the loss.

In the second stage of model training, we additionally introduce the KL divergence loss, and the loss function is combined in the following way:

$$\mathcal{L} = min \ [\lambda_{giou} \mathcal{L}_{giou}(b_v, b_v^*) + \lambda_{\mathcal{L}_1} \mathcal{L}_1(b_v, b_v^*),$$
$$\lambda_{giou} \mathcal{L}_{giou}(b_t, b_t^*) + \lambda_{\mathcal{L}_1} \mathcal{L}_1(b_t, b_t^*)]$$
$$+\lambda_{KL} \mathcal{L}_{div}(b_v, b_t). \quad (15)$$

where $\mathcal{L}_{div}$ denotes KL divergence loss, $\lambda_{KL}$ is its weight, and $b_v$, $b_t$ represent the target bounding boxes obtained from the RGB image and thermal image after modal-shared feature branches and corner predictors, respectively.

### E. Inference

The RGB-T tracking model proposed in this paper does not contain any additional post-processing operations in the inference stage, and consists of only a few steps of image sampling, forward pass, and coordinate transformation, so MACFT can run in real-time at a high speed.

## IV. EXPERIMENT AND DATA ANALYSIS

This section focuses on the experimental setup and procedure, evaluated on three major RGB-T public datasets (RGBT234 [21], LasHeR [22], and VTUAV [18]), and compared with current state-of-the-art models.

### A. Implementation Details

#### 1) Experimental platform setup

MACFT training was performed in Python 3.8.13, Pytorch 1.7.1, and CUDA 11.7. All experiments were conducted on a workstation equipped with two RTX 3090 GPUs, a Core i7 12700k CPU, and 32GB of RAM.

#### 2) Parameter details

The first stage of training trains visible and thermal modal-specific feature branches for 40 epochs, in total an 80 epoch training; 40 epochs training for the second stage, and an 80 epoch training for the third stage. Each epoch consists of 12,000 image samples, with a batch size of 64. All models are learned using the AdamW optimizer[48], with the weight decay set to $10^{-4}$ and the initial learning rate set to $10^{-5}$ for the backbone and $10^{-4}$ for the rest.

In particular, when testing the LasHeR dataset, the model is trained using LasHeR's own training set added to RGBT234, while when testing the RGBT234 dataset and VTUAV dataset, the model is trained using the full LasHeR dataset. In all training stages, $\lambda_{giou}$ and $\lambda_{\mathcal{L}_1}$ are set to 2 and 5, respectively, and in the second stage of training, $\lambda_{KL}$ is set to 800.

### B. Public dataset evaluation

#### 1) Evaluation Metrics

In this paper, we use the two most commonly used metrics to evaluate the performance of the tracking algorithm, which are the precision rate (PR) and the success rate (SR). The precision rate is calculated as the center location error (CLE) between the predicted bounding box and the ground truth, which is eventually expressed as the percentage of frames where the CLE is within a threshold (usually set to 20 pixels), and by varying the size of the threshold, a precision rate curve can be obtained. The success rate is calculated as the percentage of frames whose intersection over union (IoU) between the predicted bounding box and the ground truth is greater than the threshold value, and by setting different thresholds, a success rate curve can be obtained.

[8], C-COT [49], ECO [39], SOWP+RGB-T [50], SRDCF [51], CSR-DCF [52], KCF+RGB-T [53]) on the RGBT234 dataset and compare them with the algorithm proposed in this paper. The results show that our MACFT model achieves an precision rate of 85.7% and a success rate of 62.2% on the RGBT234 dataset, which is 5.7% and 4% better than baseline, respectively.

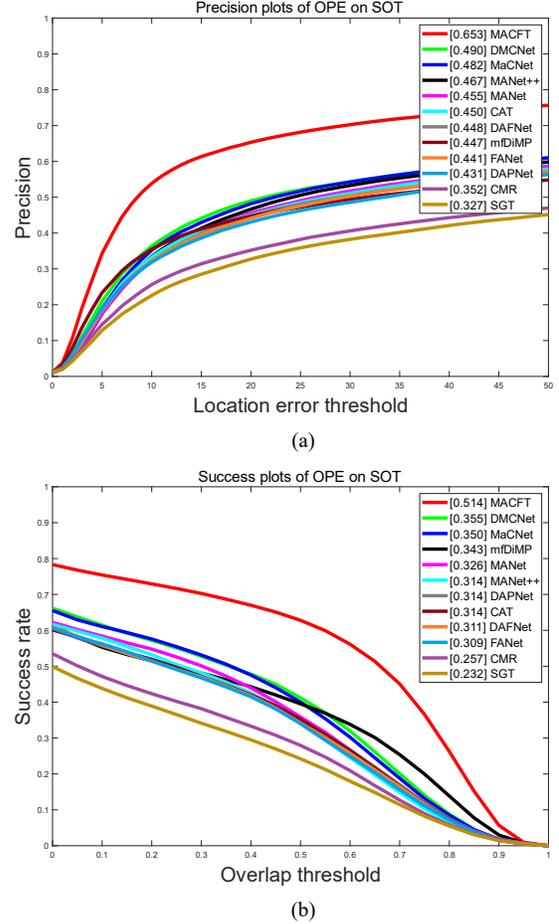

**Fig. 7.** Evaluation results on the LasHeR dataset, (a) is the precision rate curve. (b) is the success rate curve, Compared with the RGBT234 data set, the performance of each model on the two metrics has decreased significantly, indicating the challenge of the LasHeR dataset. MACFT is much better than other similar models in both metrics, which fully proves its advantages in the face of challenging scenarios.

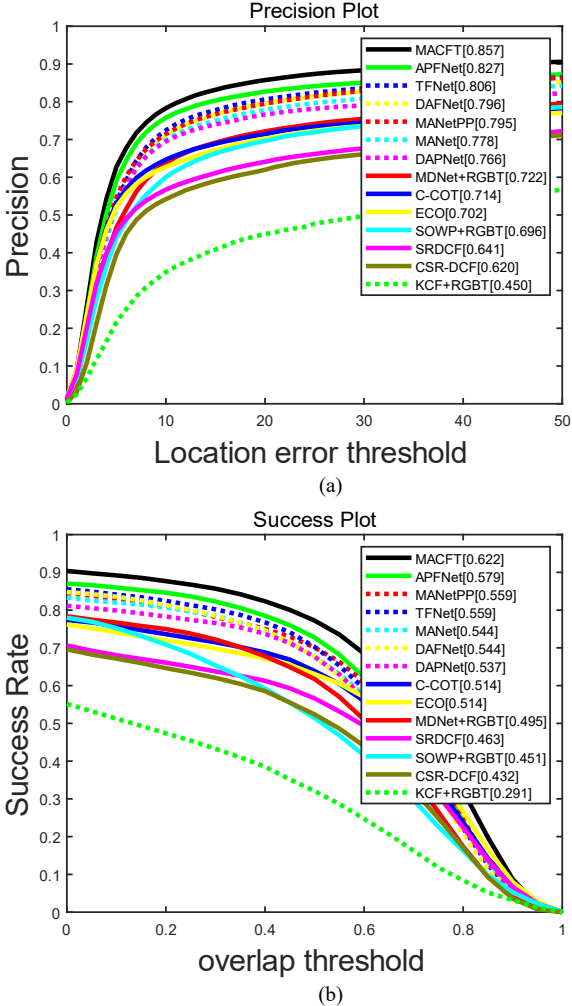

**Fig. 6.** Evaluation results on the RGBT234 dataset, (a) is the precision rate curve. (b) is the success rate curve, MACFT achieves state-of-the-art performance on both metrics.

### 2) Evaluation on RGBT234 dataset

The RGBT234 dataset contains a total of 234 RGB-T video sequences with a total of about 116.7K frames, 12 different challenge attributes, no division between the training and test sets, and the images of the two modalities are not precisely aligned. As shown in Fig. 6, we evaluate 13 different RGB-T tracking algorithms(APFNet [1], TFNet [7], DAFNet [20], DAPNet [40], MANet++ [9], MANet [10], MDNet+RGB-T

### 3) Evaluation on LasHeR dataset

The LasHeR dataset is the largest RGB-T tracking dataset with accurate annotation and alignment, with a total of about 734.8K frames, 19 challenge attributes, divided training and test sets, and a higher tracking difficulty compared to the RGBT234 dataset. To further evaluate the effectiveness of the model proposed in this paper, as shown in Fig. 7, we compare its results evaluated on the LasHeR dataset with 12 RGB-T tracking algorithms (DMCNet [16], MaCNet [3], mfDiMP [45], MANet [10], MANet++ [9], CAT [54], DAPNet [40], DAFNet [20], FANet [11], CMR [55], SGT [36]) were compared. Our MACFT model achieves 65.3% precision rate and 51.4% success rate on the LasHeR dataset, which is a 7.5% and 5% improvement respectively compared to baseline.

TABLE I
QUANTITATIVE COMPARISON AGAINST STATE-OF-THE-ART RGB-T TRACKERS
ON THE VTUAV SHORT-TERM SUBSET.

| Trackers | PR | SR | FPS |
|---|---|---|---|
| MACFT | 80.1% | 66.8% | 31.7 |
| HMFT[18] | 75.8% | 62.7% | 30.2 |
| ADRNet[4] | 62.2% | 46.6% | 25.0 |
| mfDiMP[45] | 67.3% | 55.4% | 28.0 |
| DAFNet[20] | 62.0% | 45.8% | 21.0 |
| FSRPN[56] | 65.3% | 54.4% | 30.3 |

TABLE II
QUANTITATIVE COMPARISON AGAINST STATE-OF-THE-ART RGB-T TRACKERS
ON THE VTUAV LONG-TERM SUBSET.

| Trackers | PR | SR | FPS |
|---|---|---|---|
| MACFT | 54.1% | 46.7% | 31.7 |
| HMFT-LT[18] | 53.6% | 46.1% | 8.1 |
| HMFT[18] | 41.4% | 35.5% | 30.2 |
| ADRNet[4] | 17.5% | 25.3% | 25.0 |
| mfDiMP[45] | 31.5% | 27.2% | 28.0 |
| DAFNet[20] | 25.3% | 18.8% | 21.0 |
| FSRPN[56] | 36.6% | 31.4% | 30.3 |

### 4) Evaluation on VTUAV dataset

The VTUAV dataset is a large-scale RGB-T dataset designed for unmanned aerial vehicle (UAV) tracking, and it is currently the highest resolution RGB-T dataset that considers both short-term and long-term tracking scenarios. However, the disadvantage of VTUAV is that the visible and thermal modalities are not aligned, and each frame in the sequence is not precisely manually annotated. Nevertheless, from another perspective, such data is more in line with the images obtained in real-world applications.

We evaluated our approach on both short-term and long-term tracking subsets of the VTUAV dataset and compared it with state-of-the-art RGB-T tracking algorithms. The results are shown in Table I and Table II. In particular, HMFT-LT is a model optimized specifically for long-term tracking (using GlobalTrack[57] as the global detector and RT-MDNet[25] as the tracking selector). When RT-MDNet recognizes that the target does not exist, GlobalTrack is selected to search for the target in the entire image.

In the short-term subset evaluation, MACFT achieves a precision rate of 80.1% and a success rate of 66.8%, far surpassing the second-ranked. In the long-term subset evaluation, MACFT achieves a precision rate of 54.1% and a success rate of 46.7%. It can be observed that even without specific optimization for long-term tracking tasks, MACFT still achieves state-of-the-art tracking performance on the long-term subset, which also demonstrates its potential for long-term tracking tasks.

### 5) Evaluation based on challenge attributes

To better evaluate the effectiveness of our proposed MACFT model, we evaluated it on the RGBT234 dataset based on 12 labeled challenge attributes, which are: No Occlusion (NO), Partial Occlusion (PO), Heavy Occlusion (HO), Low Illumination (LI), Low Resolution (LR), Thermal Crossover (TC), Object Deformation (DEF), Fast Motion (FM), Scale Variation (SV), Motion Blur (MB), Camera Movement (CM), and Background Clutter (BC).

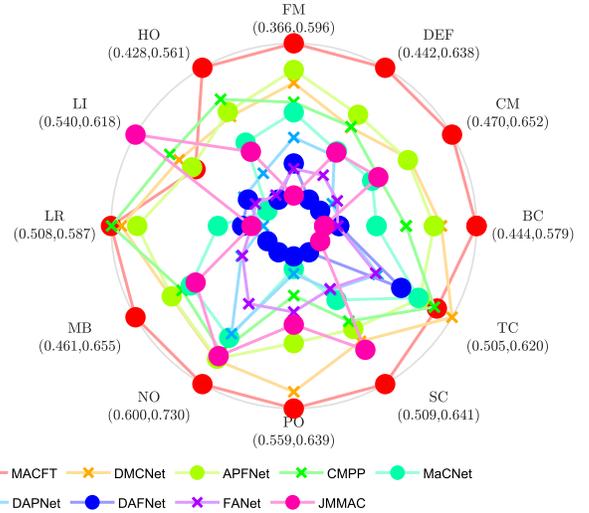

**Fig. 8.** Visualizations for different challenge attributes in the RGBT234 dataset. Taking the success rate as the evaluation metric, it can be found that MACFT has reached the optimal level in almost all challenge attributes.

As shown in Fig. 8. and Table III, our proposed model (MACFT) outperforms other state-of-the-art tracking models (respectively DMCNet[16], APFNet[1], CMPP[15], MaC-Net[3], DAPNet[40], DAFNet[20], FANet[11], JMMAC[17])in most challenging attributes, especially in scenarios involving object occlusion and deformation, further validating the effectiveness of the tracking model.

### 6) Qualitative Evaluation

Our results for the tracking performance of MACFT on several challenging video sequences are presented in Fig. 9. These sequences included object deformation, camera movement, object occlusion, low illumination, high illumination, and other challenges. To test our method, we compared it with several advanced RGBT trackers. This comparison indicates that MACFT had a higher ability to utilize the modality complementary information while excluding the interference of low-quality modality. Furthermore, MACFT had a more accurate understanding of the high-level semantic features of the target, making it more robust in tracking.

## C. Ablation Study

### 1) Model Pruning Experiments

To verify the validity of our model structure, specific anal-yses were performed for each component, and we set up several variants of the MACFT model (prefix B for baseline and DM for dual-modal):

(1) MACFT(B-T): uses the baseline model + thermal modality, (2) MACFT(B-RGB): uses the baseline model + visible modality, (3) MACFT(DM): uses 2 modal-specific branches but removes all fusion modules based on attention mechanism, only concatenating the features from the two modalities and sending them to a fully connected layer for fusion, (4) MACFT(DM+CAM): uses 2 modal-specific branches (removing modal-shared feature branch) and uses 6 cascaded CAM modules for modality fusion, (5) MACFT(DM+MAM): uses 2 modal-specific branches (removing modal-shared feature branch) and using 6 cascaded MAM modules for modality fusion, (6) MACFT(DM+CAM+COM): uses all three branches and uses CAM modules in early feature fusion, but in late feature fusion, using CAM modules instead of all MAM modules, (7) MACFT(w/o-FT): uses the complete model of MACFT but remove the finetune operation of the backbone.

TABLE III
PRECISION RATE AND SUCCESS RATE (PR/SR) OF MACFT AND OTHER TRACKERS UNDER DIFFERENT CHALLENGE ATTRIBUTE SUBSETS ON RGBT234 DATASET.
RED, BLUE, AND GREEN REPRESENT THE FIRST TO THIRD PLACE RESPECTIVELY

|      | DMCNet    | CMPP      | APFNet    | FANet     | DAPNet    | MaCNet    | DAFNet    | JMMAC     | MACFT     |
|------|-----------|-----------|-----------|-----------|-----------|-----------|-----------|-----------|-----------|
| NO   | 93.7/70.6 | 95.6/67.8 | 96.0/70.5 | 84.1/65.1 | 91.9/68.0 | 94.0/68.4 | 83.6/60.0 | 91.6/70.2 | 97.3/73.0 |
| PO   | 89.9/63.0 | 85.5/60.1 | 85.8/60.4 | 83.3/58.8 | 79.6/56.8 | 81.1/56.5 | 80.1/55.9 | 79.9/59.4 | 85.8/63.9 |
| HO   | 71.5/51.2 | 73.2/50.3 | 75.0/51.6 | 64.8/43.2 | 65.3/45.4 | 71.9/48.5 | 65.6/42.8 | 67.8/47.6 | 77.8/56.1 |
| LI   | 85.3/59.2 | 86.2/58.4 | 86.8/58.5 | 80.4/54.7 | 80.0/55.4 | 81.4/54.0 | 82.5/55.1 | 88.4/61.8 | 82.7/58.3 |
| LR   | 83.2/58.1 | 86.5/57.1 | 83.8/57.3 | 75.9/51.4 | 72.6/50.8 | 77.9/53.1 | 77.6/51.9 | 75.3/51.4 | 84.2/58.7 |
| TC   | 87.0/61.9 | 86.4/59.2 | 84.4/59.1 | 79.3/55.3 | 77.8/55.5 | 81.5/59.0 | 81.8/57.5 | 72.7/50.4 | 82.7/60.7 |
| DEF  | 75.4/54.9 | 75.0/54.1 | 79.4/56.8 | 66.4/47.8 | 70.3/50.7 | 73.5/51.4 | 65.1/44.2 | 68.0/51.2 | 83.4/63.8 |
| FM   | 81.0/53.6 | 78.6/50.8 | 86.6/55.6 | 69.1/40.6 | 70.3/45.4 | 80.9/49.2 | 65.6/41.5 | 68.0/36.6 | 85.0/59.6 |
| SV   | 82.3/59.8 | 81.1/57.8 | 83.7/58.5 | 75.0/54.6 | 76.9/54.6 | 78.3/55.7 | 73.7/50.9 | 81.0/60.6 | 86.1/64.1 |
| MB   | 79.4/60.3 | 75.4/54.1 | 80.1/60.2 | 65.6/49.8 | 66.9/49.9 | 75.8/57.4 | 62.9/46.1 | 74.3/56.7 | 86.9/65.5 |
| CM   | 81.2/59.4 | 75.6/54.1 | 81.0/59.1 | 68.2/49.4 | 67.8/48.7 | 75.9/54.2 | 68.1/47.0 | 74.2/55.0 | 88.4/65.2 |
| BC   | 82.1/54.7 | 83.2/53.8 | 81.0/54.1 | 73.2/45.5 | 68.4/45.9 | 79.6/49.0 | 78.3/45.7 | 63.8/44.3 | 85.0/57.9 |
| ALL  | 83.9/59.3 | 82.3/57.5 | 82.7/57.9 | 76.4/53.2 | 76.6/53.7 | 79.0/55.4 | 79.6/54.4 | 79.0/57.3 | 85.7/62.2 |

The performance of the above control groups on the two RGB-T datasets is shown in detail in Table III. The experimental results show that each component in MACFT improves the tracking performance to different degrees, and also indicates that the designed modal-shared branch and modality fusion network are effective.

In particular, comparition between MACFT(DM+MAM) and MACFT(DM+CAM) confirms our view that since the features from different modalities are sufficiently self-enhanced in the backbone, it does not make sense to simply add a self-attention operation after the backbone without introducing modal-shared branch. However, after introducing modal-shared feature branch and cross-fusing them with modal-specific branches, the feature distribution is changed so there is room for learning with the introduction of the self-attention operation.

### 2) Parameter Tuning Experiments

To find the optimal number of MAM modules in series under the current volume of dataset, we evaluated the precision rate and success rate on the LasHeR test set and conducted experiments on the number of MAM modules in series from 4 to 8, respectively, and the results are shown in Fig. 10. It can be found that the best results are obtained with 6 MAM modules in series under the current volume of training dataset, and using less number of MAM modules is not enough to build the relationship of complementary information interaction between modalities, while using more MAM modules makes the model parameters not adjusted properly under the same epoch of training due to the limitation of the volume of data.

### 3) Parameter Size and Inference Speed

As an object tracking algorithm, the inference speed of MACFT is crucial as it directly affects the practical usability of the algorithm. Table V shows the inference speed of MACFT and its variants on an RTX 3090 GPU, the success rate of tracking on the LasHeR dataset, and the number of trainable parameters for these models.

It can be observed that the inclusion of the modal-shared branch has a significant impact on the inference speed. If the requirement for tracking performance is not very high but the demand for inference speed is high in practical applications, a light version without the modal-shared branch can be considered.

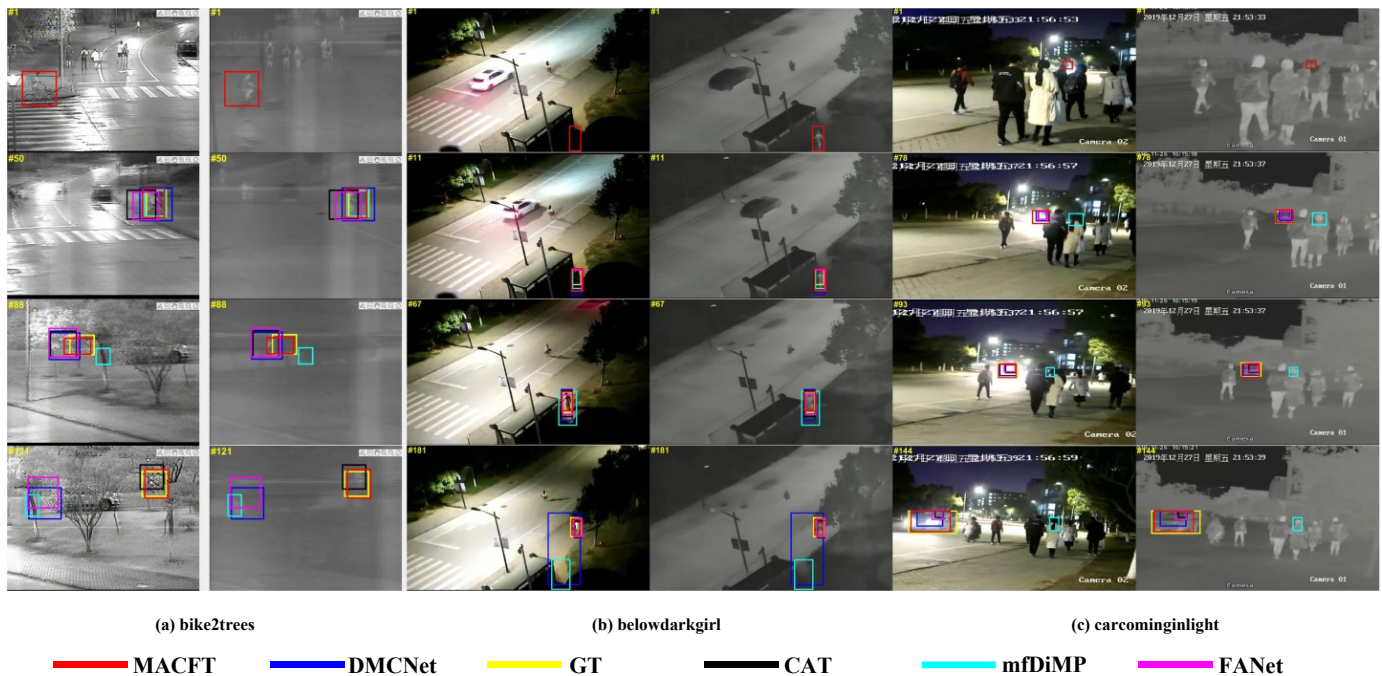

**(a) bike2trees**      **(b) belowdarkgirl**      **(c) carcominginlight**

**— MACFT**   **— DMCNet**   **— GT**   **— CAT**   **— mfDiMP**   **— FANet**

Fig. 9. A visual example of the effectiveness of our proposed tracking method. (a) (b) (c) are three challenging sequences selected from the LasHeR dataset, and also show the tracking results of other advanced RGB-T trackers for comparison.

TABLE IV
EVALUATION RESULTS OF MACFT AND ITS VARIANTS ON RGBT234 DATASET AND LASHER DATASET

| Trackers | LasHeR | | RGBT234 | |
|---|---|---|---|---|
| | PR($\downarrow$) | SR($\downarrow$) | PR($\downarrow$) | SR($\downarrow$) |
| MACFT(B-T) | 42.8%(-22.5%) | 33.2%(-18.2%) | 67.4%(-18.3%) | 48.8%(13.4%) |
| MACFT(B-RGB) | 57.8%(-7.5%) | 46.4%(-5.0%) | 80.0%(-5.7%) | 58.2%(-4.0%) |
| MACFT(DM) | 62.2%(-3.1%) | 48.7%(-2.7%) | 83.7%(-2.0%) | 58.9%(-3.3%) |
| MACFT(DM+CAM) | 63.8%(-1.5%) | 50.3%(-1.1%) | 83.7%(-2.0%) | 60.7%(-1.5%) |
| MACFT(DM+MAM) | 63.9%(-1.4%) | 50.2%(-1.2%) | 83.9%(-1.8%) | 60.7%(-1.5%) |
| MACFT(DM+CAM+COM) | 64.3%(-1.0%) | 50.7%(-0.7%) | 85.3%(-0.4%) | 61.6%(-0.6%) |
| MACFT(w/o-FT) | 64.3%(-1.0%) | 50.8%(-0.6%) | 84.0%(-1.7%) | 60.9%(-1.3%) |
| MACFT | **65.3%** | **51.4%** | **85.7%** | **62.2%** |

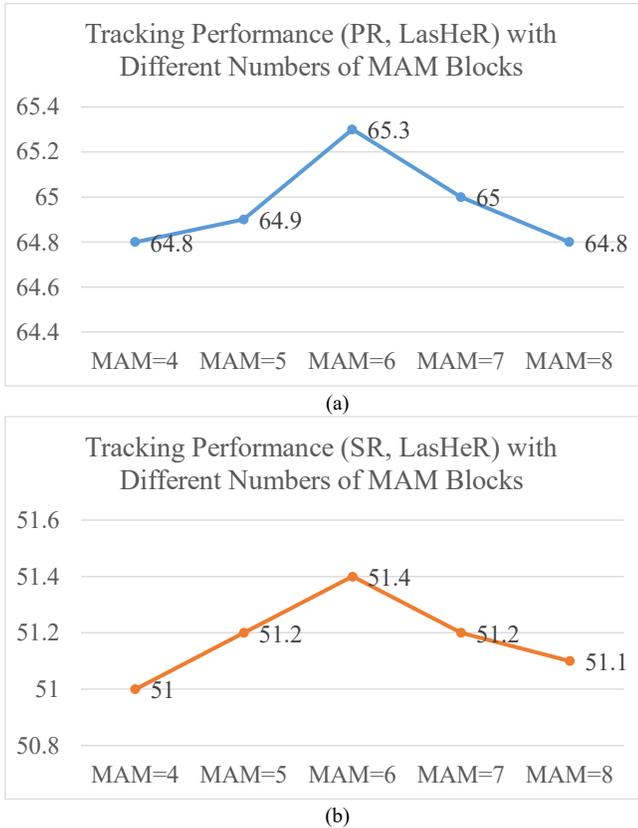

Tracking Performance (PR, LasHeR) with Different Numbers of MAM Blocks

(a)

Tracking Performance (SR, LasHeR) with Different Numbers of MAM Blocks

(b)

**Fig. 10.** The relationship between the number of MAM modules and the tracking performance, (a) indicates the precision rate, (b) indicates the success rate, both sets of data were tested on the LasHeR dataset.

TABLE V

Inference speed, success rate, and corresponding model parameters using MACFT and its variants on LasHeR
(* represents the maximum number of trainable parameters in three-stage training, excluding frozen network layers)

| Trackers | Speed (FPS) | Params* | SR |
|---|---|---|---|
| MACFT | 33.3 | 59.6M | 65.3% |
| MACFT(DM+CAM+COM) | 35.5 | 59.6M | 64.3% |
| MACFT(DM+CAM) | 50.5 | 45.4M | 63.8% |

## V. Conclusion

In this paper, we propose an RGB-T tracking method MACFT based on transformer backbone and mixed attention mechanism. Three separate branches are established to extract modal-specific features and modal-shared features, and information interaction between the template image and the search image is introduced in each layer of the backbone. In the modality fusion stage, both cross-attention and mixed attention are used to fuse the image features from different branches, which effectively reduces the noise from low-quality modality and enhances the feature information of the dominant modality.

After conducting extensive experiments, we demonstrate that MACFT can understand the high-level semantic features of the target well, thus enabling it to cope with multi-ple challenges and achieve robust tracking. Additionally, our method performs well in several public RGB-T datasets, including RGBT234, LasHeR, and VTUAV datasets. In particular, the PR/SR scores of our method improve by +7.5% + 5.0% on the testing set of LasHeR over the baseline model.